# End-to-End Machine Learning Framework for Facial AU Detection in Intensive Care Units

Subhash Nerella[a,b], Kia Khezeli[a,b], Andrea Davidson[a,c], Patrick Tighe[a,d], Azra Bihorac[a,c], Parisa Rashidi[a,b]
[a]Intelligent Critical Care Center, University of Florida, Gainesville, Florida, USA
[b]Biomedical Engineering, University of Florida, Gainesville, Florida, USA
[c]Division of Nephrology, Hypertension and Renal Transplantation, University of Florida, Gainesville, Florida, USA
[d]Department of Anesthesiology, University of Florida, Gainesville, Florida, USA

**Abstract**

**Pain is a common occurrence among patients admitted to Intensive Care Units. Pain assessment in ICU patients still remains a challenge for clinicians and ICU staff, specifically in cases of non-verbal sedated, mechanically ventilated, and intubated patients. Current manual observation-based pain assessment tools are limited by the frequency of pain observations administered and are subjective to the observer. Facial behavior is a major component in observation-based tools. Furthermore, previous literature shows the feasibility of painful facial expression detection using facial action units (AUs). However, these approaches are limited to controlled or semi-controlled environments and have never been validated in clinical settings. In this study, we present our Pain-ICU dataset, the largest dataset available targeting facial behavior analysis in the dynamic ICU environment. Our dataset comprises 76,388 patient facial image frames annotated with AUs obtained from 49 adult patients admitted to ICUs at the University of Florida Health Shands hospital. In this work, we evaluated two vision transformer models, namely ViT and SWIN, for AU detection on our Pain-ICU dataset and also external datasets. We developed a completely end-to-end AU detection pipeline with the objective of performing real-time AU detection in the ICU. The SWIN transformer Base variant achieved 0.88 F1-score and 0.85 accuracy on the held-out test partition of the Pain-ICU dataset.**

# Introduction

Patient self-report is the gold standard for pain assessment. Self-reported pain scores are commonly captured through a numeric rating scale, visual analog scale, and visual descriptor scale. These scales are subjective to the individual and have a linear representation of pain hence does not address the multidimensional aspects of pain. Beyond these, in many cases, critically ill patients cannot self-report pain for many reasons, such as being under mechanical ventilation, an altered mental state caused by the onset of delirium or dementia, and being under sedatives. In the case of non-verbal patients, ICU nurses resort to manual observation for pain assessment. Some of the nonverbal pain assessment tools employed by ICU nurses include the nonverbal pain scale (NVPS) [1], behavioral pain scale (BPS) [2], and critical care pain observation tool (CPOT) [3]. However, these assessments must be manually administered, are prone to documentation errors, and suffer significant lag between observation and documentation.

The human face plays a vital role in nonverbal communication [4, 5]. Facial behavior is a prominent factor in scoring pain using observation-based pain assessment tools such as NVPS, BPS, and CPOT. Facial expression is also characterized as a reflexive reaction to a painful stimulus or experience [6]. Most research works that looked at facial behavior used a facial anatomy-based action system referred to as the facial action coding system (FACS) [7]. The FACS system breaks down instant changes in facial expressions into individual facial action units (AUs). An AU is a contraction or relaxation of one or more facial muscles, which results in a visual appearance change on the face. Prkachin and Solomon discovered pain-related facial action units and created the Prkachin Solomon Pain Index (PSPI) score based on the FACS [8]. The PSPI score takes into consideration action units AU-4 (Brow Lowerer), AU-6 (Cheek Raiser), AU-7 (Lid Tightener), AU-9 (Nose Wrinkler), AU-10 (Upper Lip Raiser), and AU-43 (Eyes Closed), by accounting for both the presence and severity of the AUs on a (0-5) scale. The combination of AU presence and intensity results in a 16-point pain scale. The AUs must be manually coded by skilled professionals, which requires time-consuming and costly training methods and renders this strategy clinically unviable. An autonomous AU detection system can overcome the limitations of the current manual observations to document facial behavior and further facilitate real-time painful facial expression detection in the ICU.

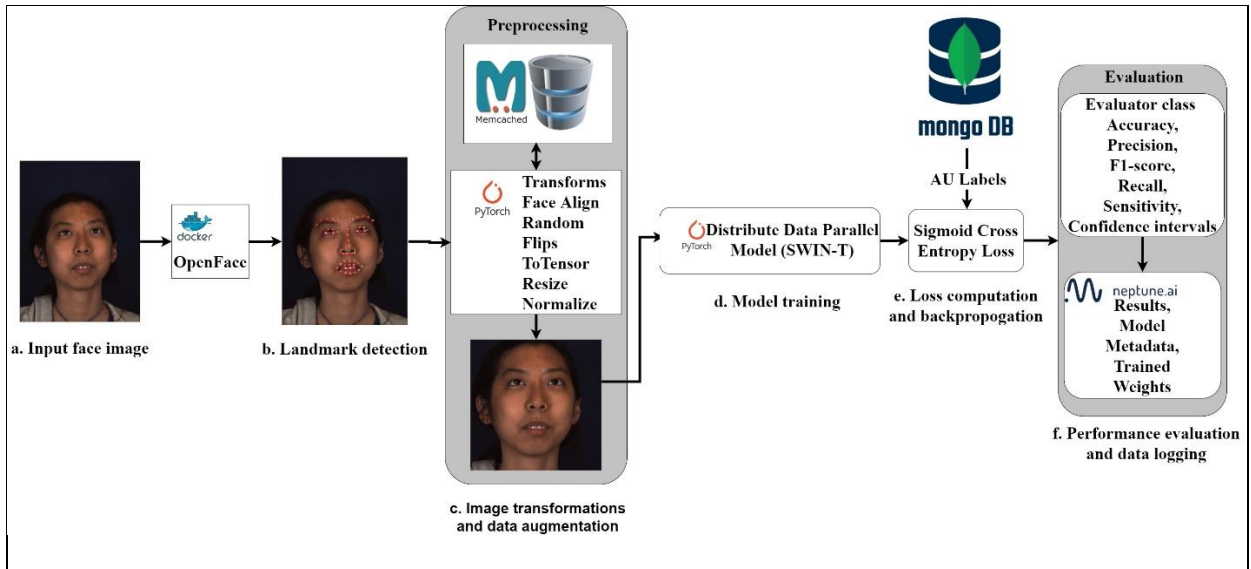

Fig 1: (a) Face image is provided as input to the Openface landmark detection tool to obtain 68 landmark locations. (b) Face image with facial landmarks (c)Image Transformations Input face image and landmarks are provided to the preprocessing block to perform face alignment and other image transformations. We use the Memcached data caching system to store the face alignment parameters to expedite the data preprocessing step. (d) The transformed image is provided as input to the Deep learning model to train. We used distributed data parallel approach to efficiently train the model on multiple GPUs. (e) All the annotation data is stored in the Mongo DB database. Labels are queried from the database to compute the loss. (f) A custom evaluator class is written to compute comprehensive model performance for all the data distributed on multiple GPUs. Finally, all the model metrics, trained weights, and metadata are uploaded to the MLops tool.

In this study, we collected data from an uncontrolled ICU environment to create Pain-ICU dataset with data compiled from critically ill adult patients. To our knowledge, the Pain-ICU dataset is the first and the largest dataset available with annotated facial AUs on videos of patients captured in a real-world ICU environment. The AUs we considered for annotating were chosen based on PSPI score and the facial features used in non-verbal pain scales.

The main contributions of this work are:

1. To the best of our knowledge, this work is the first end-to-end machine-learning pipeline (Fig 1) for AU detection in ICU environments. Our framework can be used for real-time monitoring of patients' pain levels in ICUs, which has the potential to improve the quality of patient care and reduce the burden of nurses in monitoring the patients.
2. We compare the performance of different transformer architectures and their variants on two standardized datasets for AU recognition and our Pain-ICU dataset. We demonstrate that transformer models showed on-par performance with established convolution neural networks architecture baseline with better model training and inference speed.
3. This work is the first to measure the association of different AUs with the self-reported pain scores of the patients.

## Methods

### Study Participants

The data used in this study were collected from adult patients admitted to surgical ICUs at the University of Florida Shands Health Hospital, Gainesville, Florida. The study was reviewed and approved by the University of Florida Institutional Review Board (UFIRB). We obtained informed consent from all the participants and performed data collection adhering to the regulations and guidelines of UFIRB. Table 1 shows the cohort characteristics of participants recruited for this study.

| Table 1. Patient Characteristics Table | |
|---|---|
| Participants (n = 49, number of frames =76388) | |
| Age, median (IQR) | 55 (45,64) |
| Gender, number (%) | Male 29 (59) <br> Female 20 (41) |
| Race, number (%) | White, 39 (80) <br> African American, 4 (8) <br> Other, 6 (12) |
| Length of hospital stay in days, median (IQR) | 25 (12.7,54.7) |

### Data Collection

We used a standalone cart with a camera mounted on the top to collect videos of patient's faces in the ICU. The cart is portable and can be wheeled into and out of the ICU. It is positioned in the ICU so that it does not interfere with and disrupt routine patient care in the ICU. The camera is zoomed on to the patient face to ensure we do not collect the faces of people from whom we do not have consent to record. The cart is also equipped with a computer and monitor to control the camera. We developed a graphical interface for the ICU staff and clinical coordinators to start, pause, and stop recording upon patient request or during a medical procedure.

### Data Transfer and Curation

All sensor data locally stored is encrypted, then automatically uploaded to a secured server through a secured VPN connection. Several data pipelines were developed to automate the data transfer, curation, and data preparation for annotation and model training. Docker containers were utilized to manage services and pipelines running on the server in an isolated manner. The data curation and post-processing are performed on the server. The data curation process involves removing protected health information of the patient and ensuring each patient's data is organized in the patient-specific folders on the server.

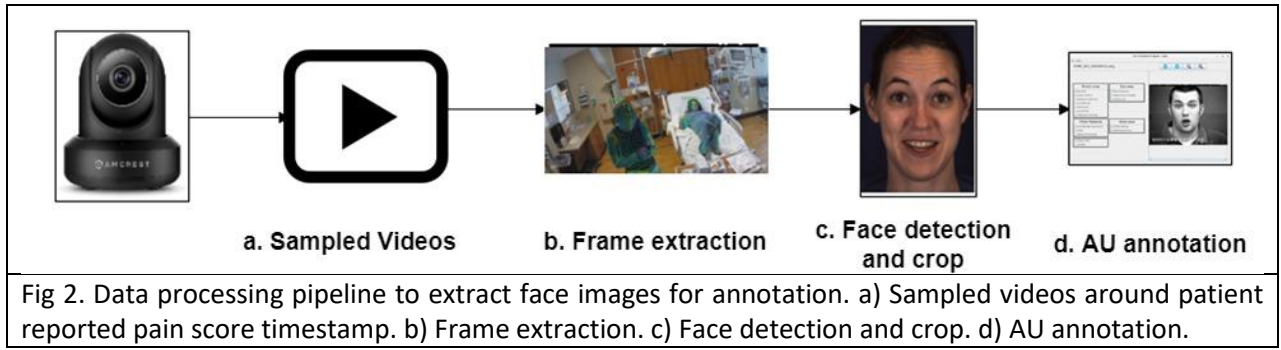

Fig 2. Data processing pipeline to extract face images for annotation. a) Sampled videos around patient reported pain score timestamp. b) Frame extraction. c) Face detection and crop. d) AU annotation.

## Data Processing

We developed a data processing pipeline to extract patient faces for AU annotation shown in fig 2. We extracted fifteen-minute videos within 1-hour proximity of patient-reported pain score timestamps. Individual Image frames are extracted from the videos using FFmpeg [11] multimedia processing tool. We used multitask cascaded convolutional network (MTCNN) [12] to detect faces in the image frame and crop them. Cropped faces were further annotated for facial AUs.

## Data Annotation

We developed an in-house secured annotation tool to annotate facial AUs in patient faces. We recruited four annotators and trained them on FACS and evaluated their performance on sample face images before they started annotating patient faces. All the annotation is performed individually by the annotators. The annotation tool is connected to a MongoDB database. In total, our Pain-ICU dataset contains 76,388 annotated frames obtained from 49 unique patients.

We use two external datasets to pretrain the deep learning models, namely BP4D and DISFAPlus datasets. BP4D dataset [9] comprises 140,000 annotated image frames with AU labels obtained from 41 participants (23 female, 18 male). The participants were asked to perform eight different tasks for emotion expression elicitation. Each image frame is annotated with AUs (1,2,4,6,7,10,12,14,15,17,23,24) and 49 facial landmarks. DISFAPlus [10] dataset is an extension to the DISFA dataset consisting of posed facial expressions from 9 subjects. Image frames are labeled for 12 FACS AUs and facial landmarks. The AUs we detected from different datasets, along with their corresponding description, are shown in Table 2.

Table 2. Facial AUs present in BP4D, DISFAPlus, and Pain-ICU Datasets

| AU | Description | BP4D[9] | DisfaPlus[10] | Pain-ICU (ours) |
|---|---|---|---|---|
| 1 | Inner Brow Raiser | ✓ | ✓ | |
| 2 | Outer Brow Raiser | ✓ | ✓ | |
| 4 | Brow Lowerer | ✓ | ✓ | ✓ |
| 5 | Upper Lid Raiser | | ✓ | |
| 6 | Cheek Raiser | ✓ | ✓ | ✓ |
| 7 | Lid Tightener | ✓ | | ✓ |
| 9 | Nose wrinkler | | ✓ | ✓ |
| 10 | Upper Lip Raiser | ✓ | | ✓ |
| 12 | Lip Corner Puller | ✓ | ✓ | ✓ |
| 14 | Dimpler | ✓ | | |
| 15 | Lip Corner Depressor | ✓ | ✓ | |
| 17 | Chin Raiser | ✓ | ✓ | |
| 20 | Lip Stretcher | | ✓ | ✓ |
| 23 | Lip Funneler | ✓ | | |
| 24 | Lip Pressor | ✓ | | ✓ |
| 25 | Lips part | | ✓ | ✓ |
| 26 | Jaw Drop | | ✓ | ✓ |
| 27 | Mouth Stretch | | | ✓ |
| 43 | Eyes Closed | | | ✓ |

## Models

JAA-Net [13], an end-to-end multitask convolutional neural network (CNN) architecture for joint learning facial AUs and facial landmarks. The JAA-net architecture comprises four different modules starting with multi-scale shared feature learning followed by face alignment, global feature learning to capture the structure and texture of the face, and adaptive attention learning for AU detection. The model performance is evaluated on multiple facial AU dataset benchmarks.

Vision Transformer (VIT) [14] was introduced as a direct application of the transformer [15] model for image recognition objective. The input image is split into fixed-size patches and transformed into linear embeddings similar to word embeddings. These patch embeddings added with positional embeddings were fed to a standard transformer encoder. An additional learnable classification token was added to the embedding sequence to perform image classification.

Swin Transformer (SWIN) [16] was proposed as a general-purpose backbone for computer vision applications. Due to the difference in the scale of the number of pixels in an image compared to words in a sentence, adapting the transformer for vision tasks becomes challenging as the self-attention mechanism of the transformer has a quadratic computational complexity with respect to the number of patches in images. To address this, SWIN utilizes shifted window-based attention, which has a linear computational complexity. Fig 3 shows the SWIN transformer architecture for the tiny variant we used for AU detection.

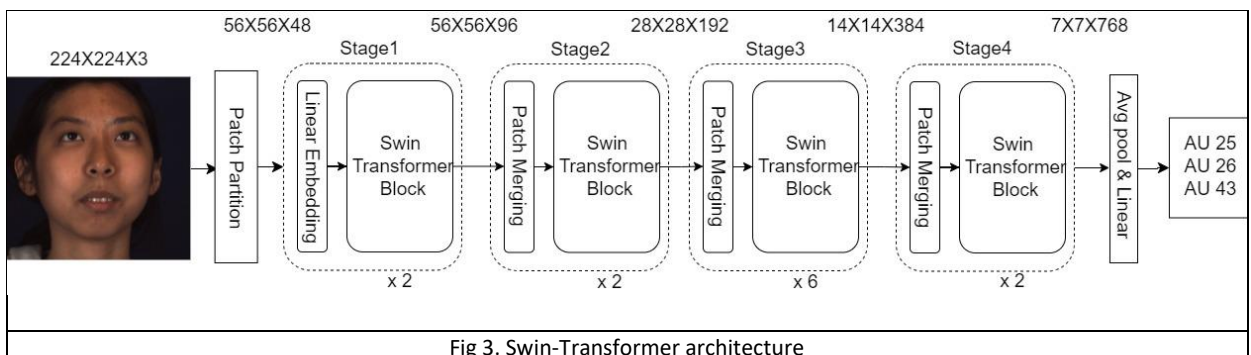

Fig 3. Swin-Transformer architecture

## Evaluation Metrics

We report the performance of models we trained and evaluated in this study using accuracy and F1-score. The model performance is computed individually for all the AUs. Hence, there is a class imbalance for each AU detection task making F1-score the appropriate measure of performance. To compute F1-score we considered True Positive (TP) as ground truth annotation specifies AU present in the image and the model detects the AU presence. True Negative (TN) when ground truth annotation specifies AU absent in a given image and the model inference is also AU absent. False Positive (FP) denotes model detects AU presence and the ground truth annotation marked AU is absent. False Negative (FN) when AU is marked as present by annotators in ground truth, but the model fails to detect AU presence. We used scikit-learn [17] library to compute the metrics, a custom evaluator class is developed to compute the aggregate performance of the model based on all the samples processed by multiple GPUs as shown in Fig 1.

## Experimental Setting

In this study, we used three different deep learning models for the objective of AU detection on our Pain-ICU dataset and external datasets BP4D and DISFAPlus. We used Pytorch [18] implementations for all the deep learning models utilized in this study. JAA-Net architecture is trained for 12 epochs with an initial learning rate of 0.01 and multiplied by 0.3 every two epochs. JAA-Net used stochastic gradient descent optimizer with a Nesterov momentum [19] of 0.9. All the hyperparameters were chosen to be the same as the original implementation. The transformer models used were trained for ten epochs using Adam [20] optimizer with a constant learning rate of 1e-5. All the transformer models are trained using PyTorch efficient multi-GPU distributed data-parallel approach shown in Fig 1 on three NVIDIA 2080 TI GPUS.

In the pipeline of end-to-end AU detection shown in Fig. 1, a raw image of the patient's face is provided as input to the OpenFace tool to detect the facial landmarks. The 68 facial landmarks are used to align and obtain a tight face crop. The affine transformation matrices used for facial alignment are stored in an in-memory database to increase the computational time complexity of the data processing step. We use random horizontal flip data transform as a data augmentation step. The images are reshaped to size 224 and normalized using the Imagenet [21] mean and standard deviation.

## Results

Facial AU detection is a challenging task in and of itself. Performing AU detection on patients in a real-world dynamic setting makes it a further arduous task. To perform AU detection in the ICU, the accuracy and speed of inference are major factors to consider. Transformer models in recent years have achieved state-of-the-art results in the computer vision domain. In this study, we used two popular vision transformer models, ViT and SWIN transformer, and trained and evaluated their performance on BP4D and DISFAPlus datasets. We compared the performance of models against JAA-Net, a state-of-the-art CNN model for AU detection. Although JAA-Net outperforms the transformer models on some AUs, on average, transformer models' performance is comparable, if not superior, to that of JAA-Net. Moreover, the run times of the transformer models are at least twice as faster (see Table 3) despite having 4X the number of trainable parameters. For these reasons, we choose the SWIN transformer to perform AU detection on our Pain-ICU dataset. We also report the model's performance on our Pain-ICU dataset when pretrained on BP4D and DISFAPlus datasets in Tables 7 and 8, respectively.

Table 3. Number of trainable parameters per model

| Model | #Params (million) | Run Duration (BP4D) |
|---|---|---|
| JAA-Net | 22.1 | ~300 minutes[*] |
| ViT-Base | 85.8 | ~90 minutes |
| SWIN-Tiny | 27.5 | ~80 minutes |
| SWIN-Base | 86.7 | ~140 minutes |

* JAA-Net implementation does not support multi-GPU training. For a fair comparison with other models, the reported value is adjusted equivalent to training on 3 GPUs.

BP4D dataset with data from 41 participants is split into three folds, similar to Zhao et al. [22]. The models are trained on two folds and evaluated the performance on the third. We used the same training and test sets for all the models to compare their performance. Table. 4 shows the performance of JAA-Net, ViT-Base, SWIN-tiny, and SWIN-BASE in terms of F1-score and accuracy on the test set.

Table 4. F1-Score and accuracy result for 12 AUs on BP4D dataset test partition.

| BP4D | F1-Score | | | | Accuracy | | | |
|---|---|---|---|---|---|---|---|---|
| AU | JAA-Net | ViT-B | Swin-T | Swin-B | JAA-NET | ViT-B | Swin-T | Swin-B |
| 1 | 0.48 | 0.45 | 0.48 | **0.55** | 0.63 | 0.67 | 0.70 | **0.75** |
| 2 | 0.45 | 0.30 | **0.51** | 0.44 | 0.74 | 0.65 | 0.75 | **0.77** |
| 4 | 0.54 | 0.46 | 0.54 | **0.55** | 0.76 | 0.78 | **0.83** | 0.78 |
| 6 | **0.81** | 0.79 | 0.76 | 0.79 | **0.81** | 0.78 | 0.78 | 0.79 |
| 7 | 0.72 | 0.71 | **0.75** | **0.75** | 0.69 | 0.69 | 0.73 | **0.75** |
| 10 | **0.86** | 0.81 | **0.86** | 0.83 | 0.80 | 0.75 | **0.82** | 0.80 |
| 12 | **0.91** | 0.88 | 0.89 | 0.90 | **0.88** | 0.85 | 0.86 | **0.88** |
| 14 | 0.56 | 0.52 | **0.60** | 0.56 | 0.59 | 0.57 | **0.60** | 0.57 |
| 15 | 0.46 | 0.32 | 0.48 | **0.50** | **0.84** | 0.77 | **0.84** | **0.84** |
| 17 | **0.63** | 0.59 | 0.50 | 0.55 | 0.71 | **0.72** | 0.67 | 0.67 |
| 23 | **0.45** | 0.33 | 0.42 | 0.40 | 0.84 | 0.81 | **0.85** | 0.75 |
| 24 | 0.36 | 0.27 | 0.33 | **0.39** | **0.85** | 0.84 | 0.83 | 0.82 |
| Avg | **0.60** | 0.54 | 0.59 | **0.60** | 0.76 | 0.74 | **0.77** | 0.76 |

DISFAPlus dataset is split into train and test splits containing data from 6 and 3 participants, respectively. Table 4 shows the performance of ViT-Base, SWIN-Tiny, and SWIN-Base models on BP4D datasets. SWIN-Tiny showed better performance compared to ViT-Base and SWIN-Base variants.

Table 5. F1-Score and accuracy result for 12 AUs on DISFAPlus dataset test partition.

| DISFAPlus | F1-Score | | | Accuracy | | |
|---|---|---|---|---|---|---|
| AU | ViT-B | Swin-T | Swin-B | ViT-B | Swin-T | Swin-B |
| 1 | 0.55 | **0.72** | 0.67 | 0.72 | **0.86** | 0.82 |
| 2 | 0.55 | **0.68** | 0.66 | 0.76 | **0.86** | 0.84 |
| 4 | 0.63 | **0.68** | 0.61 | **0.82** | **0.82** | 0.74 |
| 5 | 0.57 | 0.67 | **0.68** | 0.73 | 0.83 | **0.84** |
| 6 | 0.72 | 0.73 | **0.77** | 0.91 | 0.91 | **0.93** |
| 9 | 0.69 | 0.76 | **0.80** | 0.96 | 0.97 | **0.98** |
| 12 | **0.77** | 0.76 | 0.69 | **0.93** | 0.91 | 0.88 |
| 15 | **0.62** | 0.56 | 0.43 | **0.94** | 0.89 | 0.79 |
| 17 | 0.48 | **0.54** | 0.46 | **0.88** | 0.86 | 0.80 |
| 20 | 0.29 | **0.37** | 0.35 | **0.93** | **0.93** | 0.92 |
| 25 | **0.94** | 0.88 | 0.91 | **0.97** | 0.95 | 0.96 |
| 26 | 0.72 | **0.75** | 0.72 | **0.94** | 0.92 | 0.91 |
| Avg | 0.63 | **0.68** | 0.65 | 0.87 | **0.89** | 0.87 |

Results from table 4 and table 5 show that SWIN transformer variants outperformed other models on BP4D and DISFAPlus datasets, respectively. Therefore, we used the SWIN transformer on our Pain-ICU dataset. We split the dataset into 70-30 train test split by patients ensuring no patient data is common between train and test splits. Table 6 shows the performance of SWIN variants on the Pain-ICU dataset.

The model is trained for ten epochs. Tables 7 and 8 show the performance of SWIN variants starting with pretrained weights obtained by training on BP4D and DISFAPlus datasets, respectively.

Table 6. SWIN tiny and base variants performance on our Pain-ICU dataset test partition.

| Pain-ICU | F1-Score | | Accuracy | |
|---|---|---|---|---|
| AU | SWIN-T | SWIN-B | SWIN-T | SWIN-B |
| 25 | 0.90 | **0.91** | 0.87 | **0.88** |
| 26 | **0.89** | **0.89** | **0.89** | **0.89** |
| 43 | 0.81 | **0.85** | 0.72 | **0.79** |
| Avg | 0.87 | **0.88** | 0.83 | **0.85** |

Table 7. Performance of SWIN tiny and base variants pretrained on BP4D dataset and fine-tuned on Pain-ICU dataset test partition.

| Pain-ICU(BP4D) | F1-Score | | Accuracy | |
|---|---|---|---|---|
| AU | SWIN-T | SWIN-B | SWIN-T | SWIN-B |
| 25 | **0.88** | **0.88** | **0.85** | **0.85** |
| 26 | 0.84 | **0.86** | 0.86 | **0.87** |
| 43 | 0.75 | **0.85** | 0.67 | **0.79** |
| Avg | 0.82 | **0.86** | 0.79 | **0.84** |

Table 8. Performance of SWIN tiny and base variants pretrained on BP4D dataset and fine-tuned on Pain-ICU dataset test partition.

| Pain-ICU(DISFAPlus) | F1-Score | | Accuracy | |
|---|---|---|---|---|
| AU | SWIN-T | SWIN-B | SWIN-T | SWIN-B |
| 25 | **0.89** | **0.89** | 0.85 | **0.86** |
| 26 | **0.90** | 0.85 | **0.90** | 0.86 |
| 43 | 0.83 | **0.84** | 0.76 | **0.77** |
| Avg | **0.87** | 0.86 | **0.83** | **0.83** |

# Discussion

In this paper, we demonstrated the efficacy of attention-based vision transformer models for the facial AU detection objective. We focused on the models that have achieved state-of-the-art results on general-purpose image recognition tasks, in particular the ViT and SWIN transformers. In this study we used ViT base variant and two variants tiny and base of SWIN transformer. The basis for choosing the variants was the number of trainable parameters. Larger variants for both ViT and SWIN models have parameters on the order of ~300 million, which could result in overfitting on training data and limit the generalizability of the model. Larger models also demand more computational resources, which would adversely increase the cost of required hardware and affect the portable nature of a real-time system.

We evaluated and compared the performance of transformer variants on BP4D and DISFAPlus datasets shown in Tables 4 and 5. Table 4 also includes the JAA-Net model, a state-of-the-art CNN architecture serving as a baseline to compare the transformer models. In the case of the BP4D dataset table 4, both the SWIN-Base variant and JAA-Net achieved the best F1-score performance. SWIN-tiny achieved the best F1-score performance on the DISFAPlus dataset. In the case of both BP4D and DISFAPlus datasets, SWIN transformer variants outperformed other models. It is worth noting that the DISFAPlus dataset is of a smaller size (~57k images) compared to the BP4D dataset (~140k images). The smaller size of the dataset could be a contributing factor in SWIN tiny variant outperforming the SWIN base variant. Although JAA-Net achieved the same high performance as SWIN-Base for the BP4D dataset, it has longer training and inference time compared to transformer variants despite having fewer parameters compared to base variants of the transformer models (see Table 3). This is due to the sequential nature of JAA-Net. More specifically, JAA-Net architecture comprises multiple modules where the data is processed in sequence between some of the modules. Hence, the SWIN transformer model is more suitable for our ultimate objective, which is to perform real-time facial AU detection where inference speed is critical to achieving real-time processing speed. SWIN transformer used a unique window-based attention computation to reduce the computational complexity of attention computation. Moreover, this window-based attention computation benefited AU detection as AUs occur in localized regions of the face. Thus, the model does not need the entire context of the human face to detect whether a particular AU is present or absent. The unique localized attention computation is well-suited for the AU detection objective along with the linear computational complexity, which results in faster model inference speed than its CNN counterpart.

We trained our Pain-ICU dataset using SWIN transformer variants. Table 6 shows the performance of SWIN tiny and base variants on our Pain-ICU dataset. The base variant has achieved higher performance in terms of both F1-score and accuracy. While the base variant showed higher performance, the tiny variant performance has shown closely similar values of f1-score and accuracy. We also evaluated if starting with a pretrained model benefits the performance of models on the ICU data. Tables 7 and 8 show the performance of SWIN variants starting with pretrained weights trained on BP4D and DISFAPlus, respectively. SWIN tiny variant pretrained on DISFA achieved the same performance as that of the model without pretraining. In all other cases pretraining did not improve the model performance. There is no AU overlap between the BP4D dataset and the AUs we considered in this study from our Pain-ICU dataset, which might have resulted in a slight decrease in model performance when starting BP4D pretrained weights. AU 25 and AU 26 are common between DISFAPlus and Pain-ICU datasets. Although pretraining DISFAPlus did not increase the performance, pretraining did not result in adverse model performance.

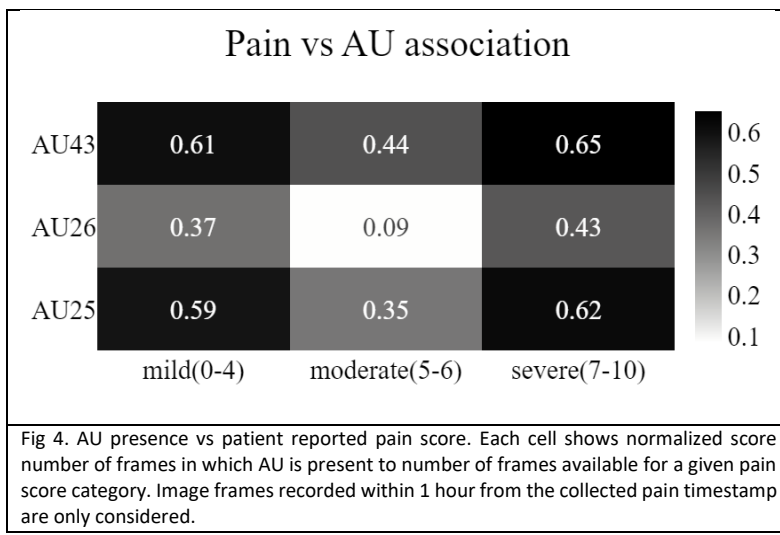

Fig 4. AU presence vs patient reported pain score. Each cell shows normalized score number of frames in which AU is present to number of frames available for a given pain score category. Image frames recorded within 1 hour from the collected pain timestamp are only considered.

Another important goal of our study is to find the association between patient pain experience and AU presence on the face. Successful identification of painful facial expressions can enable early clinical interventions for better health outcomes for patients. We are the first research group to study pain, and facial behavior association in critically ill patients admitted to ICU. To understand the association between facial AUs and pain, we used the patient self-reported pain score. We obtained the self-reported DVPRS pain score, which is measured on a 0 to 10 scale. The DVPRS pain score can be classified into three pain categories mild (0-4), moderate (5-6), and high (7-10). In Fig 4, we report AU presence against the pain category. The value shown is the percentage of frames with the AU present to the number of image frames that were collected within the 1-hour time frame of the patient-reported pain time stamp. It can be observed from fig. 4 none of the AUs are specific to mild, moderate, and high pain classes. For successful identification of painful facial expressions identifying individual AUs and AUs combination that is specific to pain is important. Although PSPI [8] score expresses pain as a combination of AUs 4,6 7,9,10, 43, it is not foolproof. In particular, a PSPI score greater than 0 does not mean pain expression because the AUs considered in PSPI are common among other facial expressions.

Our study has a few limitations. ICU is a specialized treatment space in hospitals where critically ill patients are under continuous observation. In order for our camera-mounted cart not to interfere with patient care in the ICU, it had to be positioned in an unobstructive location, which is often far away from the patient. The camera distance from the patient has impacted the image resolution in case some images. Our present work is only limited to AU 25, 26, and 43, as other AUs did not show a strong presence in the annotated images. Patients in ICU are generally under medication and sleeping, and their faces may be obstructed due to mechanical ventilation and feeding tubes, which result in no presence of AUs. It is not practical to annotate all the images during the entire patient stay. Currently, we sample videos within one hour of the patient-reported pain timestamp. To address the low AU presence problem, we are currently developing an active learning approach to prioritize the images with low presence AUs to be provided for annotation.

We have given utmost importance to patient privacy. For the same reason, we did not include any patient faces in this paper. All the face images shown in this work are from external datasets from which we obtained the license to use. Moreover, all the data used in this work is stored, processed, and analyzed on a secure server that is not connected to the internet and can only be accessed through the University of Florida health VPN. We have followed all the IRB, state, and federal rules and regulations to ensure the privacy of the patient data enrolled in our study.

## Conclusion and Future Work

In this paper, we developed a completely end-to-end deep learning framework to perform AU detection on patients admitted to ICU. We intend to use this end-to-end architecture to perform real-time facial behavior analysis in a dynamic ICU setting by deploying the model on edge devices. The top-performing model, which is the SWIN transformer, has achieved 0.88 F1-score and 0.85 accuracy on the Pain-ICU dataset test partition. The combination of end-to-end architecture and performance of the SWIN transformer showed the feasibility of a real-time system and serves as a proof of concept for our real-time framework. In the future, we intend to evaluate the model performance stratified by race and gender to ensure that our model will be fair and does not suffer from biases. As we expand our Pain-ICU dataset, we will increase the samples containing a broader range of AUs and perform a more comprehensive study of the pain and facial AU association. The successful implementation of our real-time system can improve the quality of patient care and benefit healthcare institutions in terms of cost savings and increase their efficiency.